# A Picture is Worth a Thousand (Correct) Captions: A Vision-Guided Judge-Corrector System for Multimodal Machine Translation

Team BLEU Monday

**Siddharth Betala[1], Kushan Raj[1], Vipul Betala[2], Rohan Saswade[1]**

[1]Indian Institute of Technology (IIT) Madras, [2]Independent
{ betalas5, kushan5711, vipulcbetala, rohansaswade2001 }@gmail.com
**Correspondence:** betalas5@gmail.com

## Abstract

In this paper, we describe our system under the team name *BLEU Monday* for the English-to-Indic Multimodal Translation Task at WAT 2025. We participate in the text-only translation tasks for English-Hindi, English-Bengali, English-Malayalam, and English-Odia language pairs. We present a two-stage approach that addresses quality issues in the training data through automated error detection and correction, followed by parameter-efficient model fine-tuning.

Our methodology introduces a vision-augmented judge-corrector pipeline that leverages multimodal language models to systematically identify and correct translation errors in the training data. The judge component classifies translations into three categories: correct, visually ambiguous (requiring image context), or mistranslated (poor translation quality). Identified errors are routed to specialized correctors: GPT-4o-mini regenerates captions requiring visual disambiguation, while IndicTrans2 retranslates cases with pure translation quality issues. This automated pipeline processes 28,928 training examples across four languages, correcting an average of 17.1% of captions per language.

We then apply Low-Rank Adaptation (LoRA) to fine-tune the IndicTrans2 en-indic 200M distilled model on both original and corrected datasets. Training on corrected data yields consistent improvements, with BLEU score gains of +1.30 for English-Bengali on the evaluation set (42.00 → 43.30) and +0.70 on the challenge set (44.90 → 45.60), +0.60 for English-Odia on the evaluation set (41.00 → 41.60), and +0.10 for English-Hindi on the challenge set (53.90 → 54.00).

## 1 Introduction

Machine translation (MT) for low-resource languages remains a challenging problem, particularly when dealing with multimodal data where visual context can resolve ambiguities (Specia et al., 2016; Elliott et al., 2016). The Workshop on Asian Translation (WAT) 2025 English-to-Indic Multimodal Translation Task addresses this challenge for four scheduled Indian languages: Hindi, Bengali, Malayalam, and Odia, each with distinct scripts and linguistic characteristics (Parida et al., 2019; Sen et al., 2022; Parida et al., 2024). While recent advances in neural machine translation have shown remarkable progress for high-resource language pairs (Bahdanau et al., 2014; Vaswani et al., 2017), low-resource languages continue to lag behind due to limited parallel corpora, lack of linguistic diversity in training data, and inconsistent translation quality (Sennrich and Zhang, 2019; Costa-Jussà et al., 2022).

Recent research in multimodal machine translation (MMT) has demonstrated that incorporating visual information can significantly improve translation quality, especially for ambiguous terms and culturally-specific content (Ahmed et al., 2025; Elliott et al., 2016; Calixto et al., 2017). The underlying hypothesis is that visual context provides crucial disambiguating cues that align with human cognitive processes of language understanding, which naturally rely on multiple sensory inputs (Beinborn et al., 2018). However, a critical bottleneck in training robust MMT systems for low-resource languages is the quality of parallel training data itself.

Prior work has identified systematic translation errors in the Visual Genome-based datasets used for low-resource MMT tasks (Betala and Chokshi, 2024), where captions often lack proper visual grounding, contain linguistic errors, or exhibit unnatural phrasing that can propagate through trained models. To validate these observations in the context of the WAT 2025 task, one of the authors manually evaluated a sample of the training data. This analysis confirmed numerous quality issues including mistranslations (semantic errors), visual ambiguities (terms requiring image context for disambiguation), and unnatural expressions that deviate from native speaker conventions (see Figure 2). These findings highlight a fundamental challenge: noisy training data can severely limit the effectiveness of even state-of-the-art neural MT systems.

Building on these findings, we introduce a two-stage approach that systematically addresses data quality before model training. First, we employed a vision-guided judge-corrector system powered by multimodal large language models (LLMs) to automatically identify and fix errors in training captions. Recent research has established the effectiveness of LLM-as-a-judge paradigms for quality assessment across multiple modalities (Zeng et al., 2024; Xiong et al., 2025), demonstrating their ability to provide nuanced evaluations that would traditionally require human annotators. Our judge module leverages visual context to classify each caption into one of three categories: (1) correct translations requiring no modification, (2) incorrect translations where visual context is needed to resolve ambiguities (e.g., distinguishing "dish" as food versus container), or (3) incorrect translations with poor translation quality independent of visual information (e.g., mistranslations, severe grammatical errors, or unnatural phrasing). Based on this classification, we route corrections through specialized mechanisms: a multimodal LLM (GPT-4o-mini[1](Menick et al., 2024)) regenerates captions requiring visual disambiguation, while IndicTrans2 (Gala et al., 2023), a state-of-the-art model for English-to-Indic translation, retranslates cases with pure linguistic errors. This routing strategy enables targeted correction while leveraging the strengths of each approach.

Second, we leveraged the corrected training data to fine-tune IndicTrans2 (Nair et al., 2024) using LoRA (Low-Rank Adaptation) (Hu et al., 2022), a parameter-efficient fine-tuning method (Xu et al., 2023; Han et al., 2024) that has proven effective for adapting large models to specific domains with minimal computational resources (Wong et al., 2024). To rigorously evaluate the impact of data quality on translation performance, we train separate models on both the original and corrected datasets, providing direct evidence of the benefits of automated data cleaning.

Our automated pipeline processes 28,928 training examples across four languages, correcting 19,806 captions in total. On average, 17.1% of captions per language require correction, with rates varying from 12.0% for Odia (3,486 corrections) to 24.0% for Malayalam (6,945 corrections), while Hindi and Bengali show intermediate rates of 16.3% (4,727 corrections) and 16.1% (4,648 corrections), respectively. Of the total corrections, 5,290 (26.7%) require visual context for proper disambiguation, while 14,513 (73.3%) exhibit poor translation quality addressable through retranslation. Experimental results demonstrate that training on corrected data yields consistent improvements across evaluation metrics, with notable BLEU score gains for English-Bengali (+1.30 on evaluation set, +0.70 on challenge set), English-Odia (+0.60 on evaluation set), and English-Hindi (+0.10 on challenge set). To support future research in this area, we make our corrected dataset, judge-corrector pipeline code, and trained models publicly available.[2]

Our main contributions are:

- A vision-guided judge-corrector pipeline that automatically identifies and corrects translation errors in multimodal training data through intelligent routing between visual and linguistic correction strategies

- Comprehensive analysis of error patterns in low-resource MMT datasets, processing 28,928 examples across four languages and revealing that an average of 17.1% of captions per language contain errors, with 26.7% of corrections requiring visual context for proper disambiguation

- Comparative evaluation demonstrating that LoRA finetuning on corrected data yields measurable improvements over training on original data, validating the importance of data quality in low-resource MT

## 2 Dataset and Task Description

We utilize the official datasets provided by the WAT 2025 organizers for the English-to-Indic Multimodal Translation Task. The datasets are derived from the Visual Genome corpus (Krishna et al., 2017) and consist of image-caption pairs across four target languages: Hindi (Parida et al., 2019),

---

[1] https://platform.openai.com/docs/models/gpt-4o-mini

[2] https://github.com/sid-betalol/wat-2025-english2indic-mmt

Bengali (Sen et al., 2022), Malayalam[3], and Odia[4]. Each example comprises an image, bounding box coordinates specifying a rectangular region of interest, an English caption describing that region, and corresponding translations in the target languages.

### 2.1 Task Definition

The task requires generating captions in the target language given three inputs: (1) an image, (2) a rectangular region within that image specified by bounding box coordinates, and (3) an English caption describing the visual content of that region. Participants may choose to utilize any combination of these inputs, leading to three possible translation modalities: text-only translation (using only the English caption), image-only captioning (using only the visual information), or multimodal translation (leveraging both text and image).

### 2.2 Dataset Statistics

The training set contains 28,928 examples per language, while three evaluation sets are provided for assessment: (1) a development set (Dev) with 998 examples for model validation, (2) an evaluation set (Eval) with 1,595 examples for primary assessment, and (3) a challenge set (Challenge) with 1,400 examples specifically designed to test ambiguous cases where visual context is crucial for disambiguation (Parida et al., 2024). Our official submissions were evaluated on both the Eval and Challenge sets.

| Task | Source |
| --- | --- |
| English→Hindi | Hindi Visual Genome 1.1 (Parida et al., 2019) |
| English→Bengali | Bengali Visual Genome 1.0 (Sen et al., 2022) |
| English→Malayalam | Malayalam Visual Genome 1.0[5] |
| English→Odia | Odia Visual Genome 1.0[6] |

Table 1: Tasks and their corresponding dataset sources.

Table 1 shows the data sources of datasets for each task. Table 2 shows the parallel corpus statistics across the various languages.

[3] https://lindat.mff.cuni.cz/repository/items/7ed34663-0bd4-4163-8ae9-89b2a8323269
[4] https://lindat.mff.cuni.cz/repository/items/58e6a33d-4f0f-413b-a3f3-c921e0489022
[5] https://ufal.mff.cuni.cz/malayalam-visual-genome
[6] https://ufal.mff.cuni.cz/odia-visual-genome

## 3 Methodology

The overall pipeline of our approach is shown in Figure 1.

### 3.1 Preprocessing

We perform two preprocessing steps to prepare the data for our pipeline. First, we combine the language-specific datasets into a unified format where each row contains a unique image identifier, the English caption, and corresponding translations in all four target languages.

Second, we crop the images to their specified bounding box coordinates. The dataset includes images of complete scenes along with coordinates (x, y, width, height) that define rectangular regions corresponding to each caption. We extract these regions to ensure that vision-language models focus on the precise visual content described by the captions, rather than the entire scene.

### 3.2 Manual Data Quality Assessment

Prior work by Betala and Chokshi (2024) identified systematic quality issues in Visual Genome-based datasets for multimodal machine translation, noting that captions often lack proper visual grounding, contain linguistic errors, and exhibit unnatural phrasing. These observations, made in the context of the WMT2024 English-to-Low Resource Multimodal Translation Task, highlighted a fundamental challenge: noisy training data can severely limit the effectiveness of neural MT systems, even when using state-of-the-art architectures.

Motivated by these findings, we conducted our own manual evaluation to assess whether similar issues were present in the WAT 2025 English-to-Indic datasets. One of the authors, a native Hindi speaker with formal education in Hindi through high school in the Indian education system, systematically reviewed a random sample of 1000 examples from the English-Hindi training set. This manual analysis confirmed the presence of pervasive quality issues and revealed three primary categories of errors:

**Mistranslations (Semantic Errors):** Sampled captions contained clear semantic errors where the Hindi translation did not accurately convey the meaning of the English source. These ranged from minor meaning shifts to complete mistranslations that would mislead a native speaker.

**Visual Ambiguities:** These captions contained ambiguous terms that required visual context for proper disambiguation. For example, the English word "dish" could refer to either food or a container—a distinction that native speakers would resolve by examining the accompanying image, but which was often incorrectly translated without such visual grounding.

| Set | Sentences | English | Hindi | Bengali | Malayalam | Odia |
| --- | --- | --- | --- | --- | --- | --- |
| **Train** | 28,930 | 143,164 | 145,448 | 113,978 | 107,126 | 141,652 |
| **Dev** | 998 | 4,922 | 4,978 | 3,936 | 3,619 | 4,912 |
| **Test** | 1,595 | 7,853 | 7,852 | 6,408 | 5,689 | 7,734 |
| **Challenge** | 1,400 | 8,186 | 8,639 | 6,657 | 6,044 | 8,100 |
| **Total** | 32,923 | 164,125 | 166,917 | 130,979 | 122,478 | 162,398 |

Table 2: Parallel corpus statistics (word counts) for each dataset split across different language pairs.

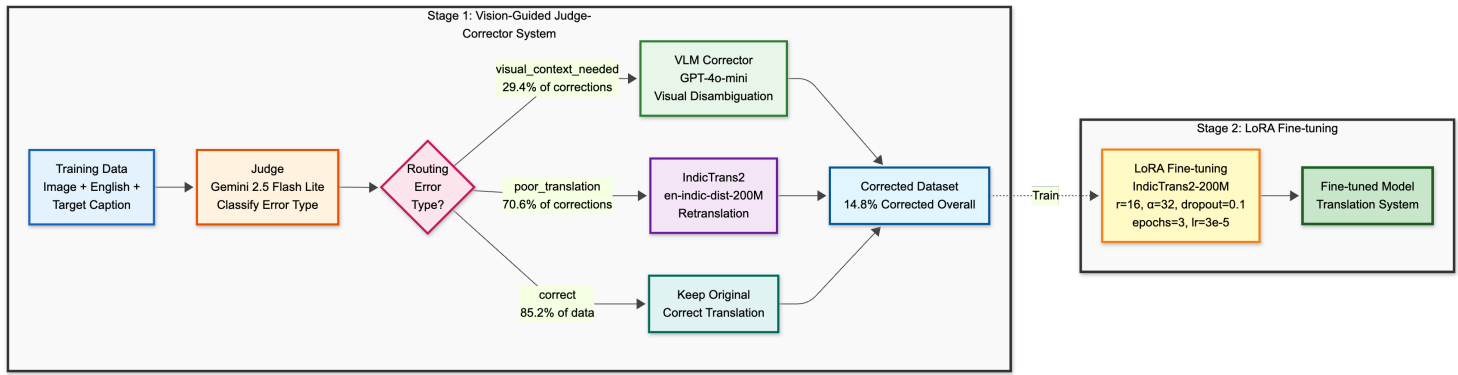

Figure 1: Overview of our two-stage approach. Stage 1 uses a vision-guided judge-corrector system to clean the training data, with 14.8% of examples corrected on average across the three languages (Hindi, Bengali, and Odia). Stage 2 applies LoRA fine-tuning to IndicTrans2.

**Unnatural Expressions:** Some captions exhibited unnatural phrasing that, while potentially understandable, deviated significantly from how native speakers would naturally express the same concept. These included awkward word choices, non-idiomatic constructions, and grammatically correct but stylistically inappropriate formulations.

It is important to note that these categories are not mutually exclusive; many captions exhibited multiple types of issues simultaneously. For instance, a single caption might contain both a mistranslation and unnatural phrasing. Detailed examples of each error category are provided in Figure 2.

This manual analysis validated the concerns raised by Betala and Chokshi (2024) in the WAT 2025 dataset context, revealing substantial quality issues across the training data. While manual correction by native speakers would be ideal, it is prohibitively expensive and time-consuming for a dataset of nearly 29,000 examples per language. These findings motivated our development of an automated judge-corrector system capable of identifying and correcting major translation errors at scale, which ultimately flagged approximately 17% of captions for correction across the four languages, focusing on cases with clear semantic errors or visual ambiguities.

### 3.3 Data Cleaning Pipeline

Our automated data cleaning pipeline employs a vision-guided judge-corrector architecture that processes each training example through three stages: judgment, routing, and correction. The system is implemented using DSPy (Khattab et al., 2024), a framework for structured prompting that enables type-safe interaction with large language models.

#### 3.3.1 Judge Module

The judge module (Table 3) evaluates each target language caption using a multimodal language model (Gemini 2.5 Flash Lite) that simultaneously analyzes the cropped image region, the English caption, and the target language translation. For each example, the judge produces four outputs:

- **Status**: Binary classification as "correct" or "incorrect"

- **Reason**: If incorrect, categorized as either "visual_context_needed" (ambiguous terms requiring visual disambiguation)

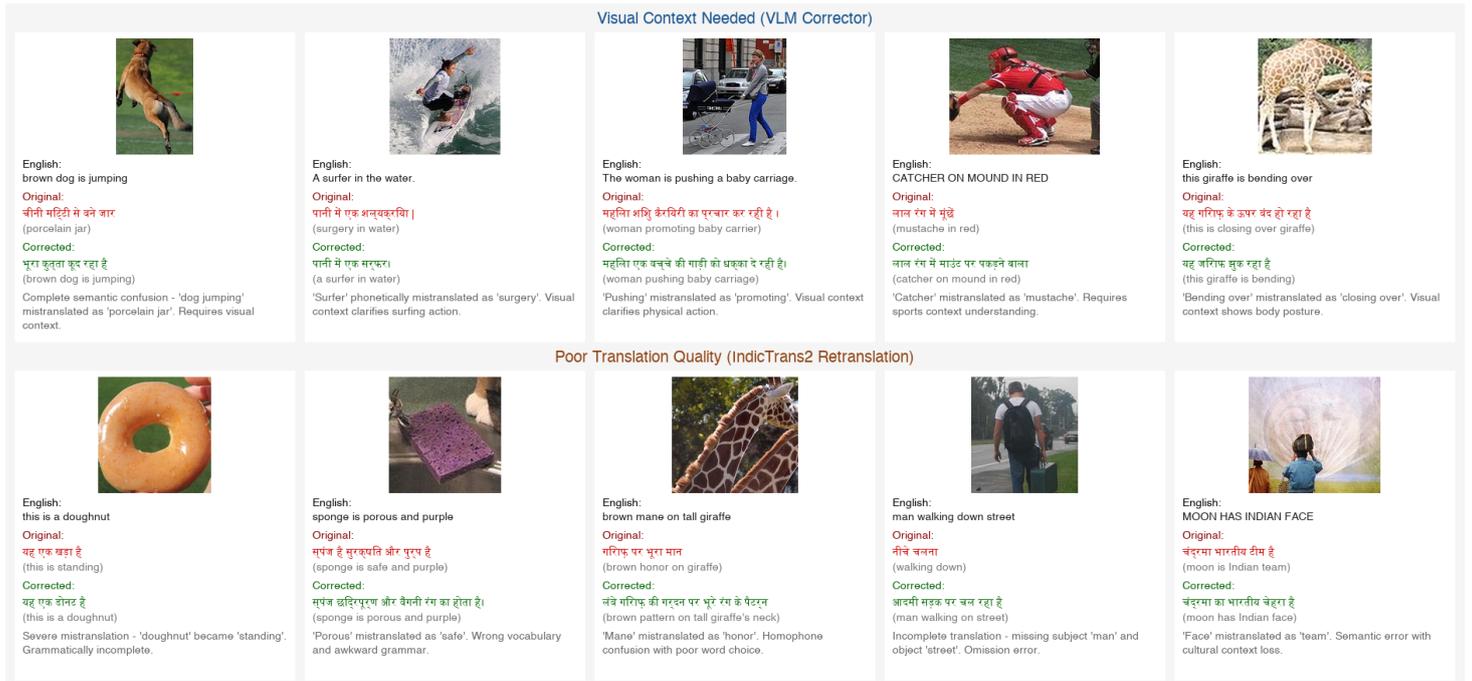

Figure 2: Training data correction examples from our judge-corrector system. Top row shows cases requiring visual disambiguation (corrected via GPT-4o-mini VLM), bottom row shows poor translation quality (corrected via IndicTrans2 retranslation). Original translations shown in red, corrections in green, with English glosses.

or "poor_translation" (linguistic errors independent of visual context)

- **Confidence**: Numerical score between 0 and 1 indicating judgment certainty
- **Explanation**: Brief justification citing the specific issue identified

The judge is explicitly instructed to focus on major issues while ignoring minor stylistic variations such as punctuation differences, optional particles, or alternative word orderings that preserve semantic equivalence. This design choice reduces false positives that would waste computational resources on unnecessary corrections while also preserving already-correct translations that could be degraded by spurious automated interventions, ensuring the pipeline focuses exclusively on substantive quality problems. Missing or empty captions are automatically classified as "visual_context_needed" without invoking the multimodal model, as they unambiguously require regeneration.

To mitigate the impact of uncertain judgments, we implement a confidence threshold: captions flagged as incorrect but with confidence below 0.7 are retained without modification. This conservative approach prevents potentially incorrect corrections in ambiguous cases where the judge's assessment may be unreliable.

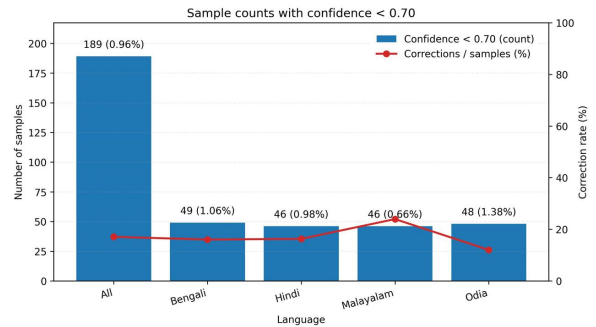

Figure 3: Judge confidence and correction statistics by language. Bars represent the count of training examples where the judge module assigned a confidence score below 0.7, resulting in retention of the original caption. The line plot shows the overall correction rate (percentage of training examples modified) for each language.

### 3.3.2 Routing and Correction

Based on the judge's classification, captions are routed through one of three paths:

**Correct captions** (∼83% of examples) are preserved without modification, maintaining the original translation quality where no issues are detected.

Instances labelled as **visual context needed** (∼27% of corrections, ∼4.5% of total examples)

> **Judge Module: Caption Quality Evaluation Prompt**
>
> **System Role:** You are an expert multilingual translator evaluating Indian language captions.
> **Primary Task:** Determine if the target language caption correctly represents what's shown in the image and accurately conveys the English caption meaning.
>
> **Focus on MAJOR issues — ignore minor stylistic differences:**
>
> **Category 1: VISUAL CONTEXT NEEDED** — Translation depends on visual information
> - Ambiguous words with multiple meanings (e.g., "dish" = food vs. container)
> - Gender-specific terms requiring visual verification
> - Spatial/directional terms (left/right/above/beside)
> - Physical attributes (color, size, material, quantity)
> - Object types/categories visible in image
>
> **Category 2: POOR TRANSLATION** — Incorrect, incomplete, or unnatural
> - Mistranslation or wrong meaning (semantic error)
> - Missing key information from English
> - Severe grammatical errors making it hard to understand
> - Completely unnatural phrasing (not just stylistic preference)
> - Wrong script or excessive script mixing
>
> **IGNORE these minor issues** — mark as CORRECT:
> - Minor punctuation differences (|, ., etc.)
> - Optional articles or particles (a/the/one equivalents)
> - Stylistic word order variations (both correct)
> - Minor postposition variations if meaning is clear
>
> **Special handling:** Empty captions → mark "incorrect" with "visual_context_needed"
>
> **Required Outputs:**
> - status: "correct" or "incorrect"
> - reason: "visual_context_needed", "poor_translation", or "none"
> - confidence: Score 0-1
> - explanation: Brief justification citing the specific issue (1-2 sentences)

Table 3: Judge module prompt template. The judge evaluates caption quality using visual context and classifies captions into three categories: correct, requiring visual context for disambiguation, or poor translation quality. Explicit instructions guide the model to focus on major issues while ignoring minor stylistic variations.

are processed by GPT-4o-mini, a multimodal language model that regenerates the caption by analyzing both the cropped image and the English source. This approach is specifically designed for cases where ambiguous terms require visual grounding for proper disambiguation. The model is provided with the original (potentially incorrect) caption for reference but is instructed to prioritize visual evidence when generating the corrected version. The prompt for this module is highlighted in Table 4.

**Poor translation** cases (∼73% of corrections, ∼12.5% of total examples) are retranslated using IndicTrans2 (Gala et al., 2023), a state-of-the-art neural machine translation model specifically trained for English-to-Indic language pairs. This routing strategy leverages IndicTrans2's strong performance on pure translation tasks while reserving the more expensive multimodal LLM for cases where visual context is essential.

### 3.3.3 Implementation Details

The pipeline processes all four target languages (Hindi, Bengali, Malayalam, Odia) concurrently for each image, with rate limiting to manage API costs and comply with provider constraints.

The pipeline processes all four target languages concurrently for each image, with rate limiting (maximum 4 concurrent API calls) to manage costs and comply with provider constraints. To optimize efficiency, images are loaded once per example and reused across all language evaluations, while automatic checkpointing every 100 examples enables recovery from interruptions.

Based on the parallel corpus statistics as shown in Table 2, the corrector module receives explicit guidance on typical caption lengths for each target language to ensure natural output: Hindi and Odia captions should match English word counts, while Bengali should be approximately 20% shorter and Malayalam 25% shorter. We speculate that these language-specific guidelines might help maintain

> **Corrector Module: Natural Caption Generation Prompt**
>
> **System Role:** Expert translator creating natural Indian language captions.
> **Generation Process:**
> 1. **Analyze the IMAGE** to understand visual context
> 2. **Use visual details** to resolve ambiguities (e.g., "dish" = food vs. container)
> 3. **Create natural captions** that native speakers would use
> 4. **Match English meaning** while respecting target language style
>
> **Target Language Length Guidelines** (be concise, not verbose):
> - **Hindi**: Similar word count to English
> - **Bengali**: ∼20% fewer words than English
> - **Malayalam**: ∼25% fewer words than English
> - **Odia**: Similar word count to English
>
> **Important Note:** Original caption may be wrong/missing — **trust the IMAGE first**
>
> **Required Outputs:**
> - `corrected_caption`: Natural, accurate caption in target language
> - `explanation`: What you corrected and why (1-2 sentences)

Table 4: Corrector module prompt template. The corrector generates natural captions using visual evidence with language-specific length guidelines to ensure native-like output. The model is instructed to prioritize image information when the original caption may be incorrect or missing.

stylistic consistency with native speaker conventions while preventing unnecessarily verbose or overly terse translations.

### 3.4 Model Finetuning

Following data cleaning, we fine-tune the Indic-Trans2 en-indic 200M distilled model[7] (Gala et al., 2023) using Low-Rank Adaptation (LoRA) (Hu et al., 2022), a parameter-efficient fine-tuning technique that updates only a small subset of model parameters while keeping the base model frozen.

#### 3.4.1 Data Preparation

We prepare the cleaned training data for multilingual fine-tuning by converting it into a unified format suitable for Indic-Trans2. Each training example consists of four components: (1) the English source text (`english_caption`), (2) the target language translation (either `{language}_corrected` or `{language}_original` depending on the training configuration), (3) the source language code (`eng_Latn`), and (4) the target language code in FLORES-200 format (Costa-Jussà et al., 2022; nll, 2024) (`hin_Deva` for Hindi, `ben_Beng` for Bengali, `mal_Mlym` for Malayalam, `ory_Orya` for Odia).

The training data is processed through the official IndicTransToolkit processor (Gala et al.,

---
[7] https://huggingface.co/ai4bharat/indictrans2-en-indic-dist-200M

2023), which applies language-specific preprocessing including script normalization and tokenization conventions. We create one training example per language per image, resulting in 115,712 total training examples (28,928 images × 4 languages). The development set follows the same preprocessing pipeline, using the original `{language}_text` columns from the official development split.

#### 3.4.2 LoRA Configuration

We apply LoRA to the attention projection layers (`q_proj` and `v_proj`) of the transformer encoder-decoder architecture (Vaswani et al., 2017). The LoRA configuration uses rank $r = 16$ with scaling factor $\alpha = 32$, resulting in approximately 0.8M trainable parameters compared to the base model's 200M parameters (0.4% of total parameters). We set the LoRA dropout probability to 0.1 to prevent overfitting on the relatively small training set. This parameter-efficient approach enables training on consumer hardware while maintaining competitive performance (Hu et al., 2022).

#### 3.4.3 Training Configuration

Training is conducted using the Hugging Face Transformers library (Wolf et al., 2020) with the Seq2SeqTrainer class. We use the following hyperparameters: per-device batch size of 8 with 2-way gradient accumulation, resulting in an effective global batch size of 32 across 2 GPUs; learning rate of $3 \times 10^{-5}$ with 500 warmup steps using linear scheduling; weight decay of 0.01; and maximum gradient norm of 1.0 for stability. We train for 3 epochs, which balances convergence with computational efficiency. All experiments use float32 precision to ensure numerical stability across different hardware platforms.

The model is trained in a **multilingual** fashion, where a single model learns to translate from English to all four target languages simultaneously (Aharoni et al., 2019). Each training batch contains examples from all languages, enabling the model to share representations across related Indic languages while learning language-specific translation patterns through the FLORES-200 language codes that prefix each input. This multilingual approach has been shown to improve performance for lower-resource languages through cross-lingual transfer (Arivazhagan et al., 2019).

The training employs standard sequence-to-sequence preprocessing: input sequences are tokenized using the IndicTrans2 tokenizer with a maximum length of 256 tokens, and the DataCollatorForSeq2Seq applies padding to create uniform batch sizes while masking padding tokens in label sequences with -100 to exclude them from loss computation. During inference, we use greedy decoding (beam size 1, 'num_beams=1') as a workaround for known beam search compatibility issues in the IndicTrans2 implementation.

### 3.4.4 Inference

For inference, we load the trained LoRA adapters and merge them with the base IndicTrans2 model using PEFT's `merge_and_unload()` method (Mangrulkar et al., 2022), eliminating the overhead of adapter routing during generation. Translations are generated using the IndicTransToolkit's preprocessing and postprocessing pipelines to ensure consistency with the model's training format. We translate the evaluation and challenge sets in batches of 16 with a maximum generation length of 256 tokens.

### 3.4.5 Submitted Systems

Due to resource and time constraints, we submit results for three language pairs: English-Hindi, English-Bengali, and English-Odia. We do not submit results for English-Malayalam.

To rigorously evaluate the impact of data cleaning on translation quality, we submit translations from two systems: (1) a LoRA model trained on the original (uncorrected) training data, and (2) a LoRA model trained on our corrected training data. Both models use identical architectures, hyperparameters, and training procedures, with the only difference being the quality of the training captions. This controlled comparison allows us to directly attribute performance differences to the data cleaning pipeline.

### 3.5 System Classification

We classify our submissions according to the WAT 2025 task guidelines across four dimensions, as specified by the task organizers (Parida et al., 2024).

First, we participate in the **unconstrained track** due to our use of multimodal large language models, specifically GPT-4o-mini[8] (Menick et al., 2024) and Gemini 2.5 Flash Lite[9] (Comanici et al., 2025), as part of our automated data cleaning pipeline. While these models are not used during inference, their use in training data preparation exceeds the pretrained model restrictions of the constrained track.

Second, our approach is classified as **text-only translation**. Although our data cleaning pipeline leverages visual information to identify and correct translation errors in the training set, the final trained model does not use images during inference. At translation time, the model receives only the English source text as input, without access to the corresponding image or bounding box information.

Third, we are **domain-unaware**, using exclusively the officially provided training data (28,928 examples per language) without incorporating the full English Visual Genome corpus or any additional external datasets. Our data cleaning process operates only on the provided parallel captions, improving their quality without introducing new training examples.

Finally, our system is **multilingual**, employing a single IndicTrans2 model that simultaneously translates from English to all four target languages (Hindi, Bengali, Malayalam, and Odia). Rather than training separate pairwise models for each language pair, our multilingual approach enables cross-lingual transfer and parameter sharing across the related Indic languages (Arivazhagan et al., 2019), while using FLORES-200 language codes to distinguish target languages during generation.

---

[8] https://platform.openai.com/docs/models/gpt-4o-mini
[9] https://deepmind.google/models/gemini/flash-lite/

# 4 Results

We present the results of our two-stage approach: first analyzing the impact of our automated data cleaning pipeline on training data quality, then evaluating how these improvements translate to translation performance on the official evaluation and challenge test sets.

## 4.1 Data Cleaning Statistics

Our automated judge-corrector pipeline processed all 28,928 training examples across four target languages, identifying and correcting quality issues in 19,806 captions (17.1% of total examples). Table 5 summarizes the correction statistics by language.

The correction rates vary significantly across languages, with Malayalam requiring the most corrections (24.0%) and Odia requiring the fewest (12.1%). This variation likely reflects differences in the original annotation processes for each language dataset (Parida et al., 2019; Sen et al., 2022). Across all languages, the majority of corrections (73.3%, 14,513 out of 19,806) address poor translation quality that can be resolved without visual context, while 26.7% (5,290 corrections) involve visually ambiguous terms requiring multimodal understanding. The low number of missing captions (3 total) indicates that the original datasets were largely complete, with quality issues primarily manifesting as incorrect or unnatural translations rather than absent captions.

## 4.2 Translation Performance

Table 6 presents the main results comparing models trained on original versus corrected data across three language pairs on both the evaluation set (1,595 examples) and challenge set (1,400 examples). We report BLEU (Papineni et al., 2002) and RIBES (Isozaki et al., 2010) scores, two standard automatic metrics for evaluating machine translation quality.

### 4.2.1 Impact of Data Cleaning

The results demonstrate that data cleaning yields consistent improvements for English-Bengali across both test sets, with BLEU score gains of +1.30 on the evaluation set (42.00 → 43.30) and +0.70 on the challenge set (44.90 → 45.60). These improvements are substantial given that Bengali had a moderate correction rate (16.1%) in the training data. The challenge set improvements are particularly noteworthy, as this set specifically targets ambiguous cases where visual context is crucial for disambiguation—precisely the type of errors our vision-guided corrector is designed to address.

For English-Odia, we observe a +0.60 BLEU improvement on the evaluation set (41.00 → 41.60), though performance on the challenge set shows a marginal decline of -0.10 points. This mixed result is notable given that Odia had the lowest correction rate (12.1%) among all submitted languages, suggesting that the original Odia training data was already of relatively high quality. The slight performance decrease on the challenge set may indicate that automated correction can occasionally degrade high-quality original translations when the error rate is already low.

English-Hindi shows the smallest improvements, with identical BLEU scores (42.10) on the evaluation set and only +0.10 improvement on the challenge set (53.90 → 54.00). However, we observe consistent RIBES improvements for Hindi on the evaluation set (+0.0021), indicating better word ordering despite similar BLEU scores. The minimal BLEU improvements for Hindi may reflect that this widely-studied language already had relatively clean training data (16.3% correction rate), limiting the potential gains from automated cleaning.

### 4.2.2 Error Type Analysis

Examining the relationship between correction types and performance gains reveals instructive patterns. Bengali, which showed the largest improvements, had a balanced distribution of error types (31% visual context, 69% poor translation). This suggests that both the vision-guided corrections (handled by GPT-4o-mini) and the text-based retranslations (handled by IndicTrans2) contributed meaningfully to improved model quality. The fact that substantial improvements were achieved despite correcting only 16.1% of the training data underscores the importance of targeting high-impact errors rather than achieving perfect coverage.

The challenge set results provide evidence for the value of vision-guided corrections, particularly for Bengali which showed consistent gains across both test sets. However, Odia's slight decline on the challenge set highlights an important limitation: automated correction, even with multimodal guidance, cannot perfectly replicate human judgment and may occasionally introduce errors when applied to already high-quality translations. This sug-gests that automated cleaning provides the greatest benefit for datasets with moderate to high error rates, while datasets with very low error rates (such as Odia at 12.1%) may see diminishing or mixed returns.

| Language | Correct | Corrected | Visual | Translation |
|---|---|---|---|---|
| Hindi | 24,201 (83.7%) | 4,727 (16.3%) | 1,314 | 3,412 |
| Bengali | 24,280 (83.9%) | 4,648 (16.1%) | 1,436 | 3,211 |
| Malayalam | 21,983 (76.0%) | 6,945 (24.0%) | 1,507 | 5,438 |
| Odia | 25,442 (87.9%) | 3,486 (12.1%) | 1,033 | 2,452 |
| **Total** | **95,906 (83.0%)** | **19,806 (17.1%)** | **5,290** | **14,513** |

Table 5: Data cleaning statistics across four languages. "Visual" indicates corrections requiring visual context (handled by GPT-4o-mini), while "Translation" indicates poor translations (handled by IndicTrans2). Percentages show proportion of total 28,928 examples per language.

| Language Pair | Model | Test Set | BLEU | RIBES | Δ BLEU |
|---|---|---|---|---|---|
| English-Hindi | Original | Evaluation | 42.10 | 0.8148 | — |
| | Corrected | Evaluation | **42.10** | **0.8169** | +0.00 |
| | Original | Challenge | 53.90 | **0.8668** | — |
| | Corrected | Challenge | **54.00** | 0.8648 | +0.10 |
| English-Bengali | Original | Evaluation | 42.00 | 0.7590 | — |
| | Corrected | Evaluation | **43.30** | **0.7704** | +1.30 |
| | Original | Challenge | 44.90 | **0.8126** | — |
| | Corrected | Challenge | **45.60** | 0.8089 | +0.70 |
| English-Odia | Original | Evaluation | 41.00 | **0.8466** | — |
| | Corrected | Evaluation | **41.60** | 0.8459 | +0.60 |
| | Original | Challenge | **40.10** | **0.8727** | — |
| | Corrected | Challenge | 40.00 | 0.8703 | -0.10 |

Table 6: Translation performance comparing LoRA finetuning on original versus corrected training data. Bold indicates best performance for each language pair and test set. Δ BLEU shows the improvement (+) or degradation (-) from data correction.

### 4.2.3 Comparison to Full Finetuning Approaches

Our LoRA-based approach represents a parameter-efficient alternative to full finetuning, enabling rapid experimentation and comparison between original and corrected training data. While our results demonstrate clear benefits from data cleaning using LoRA, we note that full finetuning of IndicTrans2 could potentially yield even stronger performance. The IITP-AI-NLP-ML team, which achieved top rankings on multiple leaderboards in this shared task, employed full finetuning of IndicTrans2 across all language pairs. This suggests that the improvements we observe from data cleaning with LoRA likely represent a lower bound on the potential gains, and that combining our data cleaning approach with full finetuning could yield further performance improvements.

### 4.3 Limitations and Future Work

**Test set quality.** An important limitation of our evaluation is that we applied data cleaning only to the training set. Since the evaluation and challenge test sets were curated using the same annotation process as the training data, they likely contain similar quality issues—mistranslations, visual ambiguities, and unnatural expressions. The presence of errors in the reference translations could artificially suppress our reported BLEU and RIBES scores, as these metrics penalize deviations from the references even when our model's output may be more accurate or natural than the reference itself. If the test set references were corrected using our pipeline or through manual annotation by native speakers, the true performance of our corrected-data model would likely be higher, and the gap between original and corrected models would be more pronounced. This represents an important direction for future work: applying our data cleaning methodology to create higher-quality evaluation benchmarks for low-resource multimodal translation.

**Language coverage.** Due to resource and time constraints, we submitted results for only three of the four target languages (Hindi, Bengali, and Odia), omitting Malayalam. Given that Malayalam exhibited the highest correction rate (24.0%) in our data cleaning analysis, it represents a particularly interesting case for future investigation. The substantial number of corrections in Malayalam suggests that this language pair could benefit significantly from our approach, and we encourage future work to validate this hypothesis.

**Model capacity.** Our experiments focused exclusively on parameter-efficient LoRA finetuning rather than full model finetuning. While this enabled rapid experimentation and fair comparison between original and corrected data, it likely underestimates the full potential of our data cleaning approach. Combining our corrected training data with full finetuning could yield additional performance gains, as evidenced by the strong results achieved by teams employing full finetuning strategies.

### 4.4 Key Takeaways

Our experimental results validate three main findings:

**(1) Data quality significantly impacts translation performance:** Across three language pairs, training on corrected data yields consistent improvements or competitive performance compared to original data, with Bengali showing substantial gains (+1.30 BLEU on evaluation, +0.70 on challenge). This demonstrates that automated data cleaning can meaningfully improve translation quality for low-resource languages, even when correcting a relatively small proportion (16-17%) of the training data.

**(2) Correction effectiveness varies by initial data quality:** Languages with moderate error rates (Bengali: 16.1%) and balanced error distributions benefit most from automated correction, while languages with very low error rates (Odia: 12.1%) show more modest or mixed improvements. This suggests that automated cleaning provides the greatest value for datasets with known quality issues, and that careful analysis of error rates should guide the decision to apply automated correction.

**(3) Vision-guided correction addresses a real need:** The improvements on the challenge set—specifically designed to test ambiguous cases requiring visual context—validate the core hypothesis that multimodal language models can effectively resolve translation ambiguities that text-only approaches miss. However, the mixed results on some language pairs indicate that automated multimodal correction works best when applied to datasets with moderate error rates rather than already high-quality data. The success on Bengali (+0.70 BLEU on challenge set) demonstrates that when error rates justify intervention, vision-guided correction provides measurable value.

## 5 Conclusion

We presented a vision-guided judge-corrector system that addresses training data quality in low-resource multimodal translation. Our automated pipeline processed 28,928 examples across four languages, correcting 17.1% of captions through intelligent routing between multimodal LLMs (for visual ambiguities) and IndicTrans2 (for translation errors). LoRA finetuning on corrected data yields measurable BLEU improvements: +1.30 for Bengali (eval), +0.70 (challenge), +0.60 for Odia (eval), and +0.10 for Hindi (challenge).

Our approach demonstrates that automated data cleaning can meaningfully improve low-resource MT, particularly for datasets with moderate error rates. However, important limitations remain: test set quality issues likely suppress reported scores, automated correction cannot perfectly replicate human judgment (as seen in Odia's mixed results), and our LoRA-only experiments likely underestimate the full potential when combined with full finetuning.

Future work should focus on three key directions: (1) human evaluation by native speakers to validate improvements beyond automatic metrics, (2) applying our pipeline to create higher-quality test (eval and challenge) sets for more accurate evaluation, and (3) combining corrected data with full model finetuning to validate whether quality improvements compound with increased capacity.

We hope that our publicly released dataset, code, and models provide a foundation for future research in automated quality assurance for multimodal datasets, potentially enabling more robust and equitable AI systems across diverse languages.